\documentclass{article}
\usepackage{iclr2027_conference,times}
\usepackage{amsmath,amsfonts,bm}

\def\eqref#1{equation~\ref{#1}}
\def\1{\bm{1}}

\DeclareMathAlphabet{\mathsfit}{\encodingdefault}{\sfdefault}{m}{sl}
\SetMathAlphabet{\mathsfit}{bold}{\encodingdefault}{\sfdefault}{bx}{n}

\ifdefined\XeTeXversion\else
  \usepackage[utf8]{inputenc}
  \usepackage[T1]{fontenc}
\fi
\ifdefined\XeTeXversion
  \DeclareFontFamily{TU}{ptm}{}
  \DeclareFontShape{TU}{ptm}{m}{n}{<->"[.latex-env/fonts/LiberationSerif-Regular.ttf]:mapping=tex-text;"}{}
  \DeclareFontShape{TU}{ptm}{m}{it}{<->"[.latex-env/fonts/LiberationSerif-Italic.ttf]:mapping=tex-text;"}{}
  \DeclareFontShape{TU}{ptm}{m}{sl}{<->"[.latex-env/fonts/LiberationSerif-Italic.ttf]:mapping=tex-text;"}{}
  \DeclareFontShape{TU}{ptm}{m}{sc}{<->"[.latex-env/fonts/LiberationSerif-Regular.ttf]:mapping=tex-text;"}{}
  \DeclareFontShape{TU}{ptm}{b}{n}{<->"[.latex-env/fonts/LiberationSerif-Bold.ttf]:mapping=tex-text;"}{}
  \DeclareFontShape{TU}{ptm}{b}{it}{<->"[.latex-env/fonts/LiberationSerif-BoldItalic.ttf]:mapping=tex-text;"}{}
  \DeclareFontShape{TU}{ptm}{b}{sl}{<->"[.latex-env/fonts/LiberationSerif-BoldItalic.ttf]:mapping=tex-text;"}{}
  \DeclareFontShape{TU}{ptm}{b}{sc}{<->"[.latex-env/fonts/LiberationSerif-Bold.ttf]:mapping=tex-text;"}{}
  \DeclareFontShape{TU}{ptm}{bx}{n}{<->"[.latex-env/fonts/LiberationSerif-Bold.ttf]:mapping=tex-text;"}{}
  \DeclareFontShape{TU}{ptm}{bx}{it}{<->"[.latex-env/fonts/LiberationSerif-BoldItalic.ttf]:mapping=tex-text;"}{}
  \DeclareFontShape{TU}{ptm}{bx}{sl}{<->"[.latex-env/fonts/LiberationSerif-BoldItalic.ttf]:mapping=tex-text;"}{}
  \DeclareFontShape{TU}{ptm}{bx}{sc}{<->"[.latex-env/fonts/LiberationSerif-Bold.ttf]:mapping=tex-text;"}{}
  \DeclareFontFamily{TU}{pcr}{}
  \DeclareFontShape{TU}{pcr}{m}{n}{<->"[.latex-env/fonts/LiberationMono-Regular.ttf]:mapping=tex-text;"}{}
  \DeclareFontShape{TU}{pcr}{m}{it}{<->"[.latex-env/fonts/LiberationMono-Italic.ttf]:mapping=tex-text;"}{}
  \DeclareFontShape{TU}{pcr}{m}{sl}{<->"[.latex-env/fonts/LiberationMono-Italic.ttf]:mapping=tex-text;"}{}
  \DeclareFontShape{TU}{pcr}{b}{n}{<->"[.latex-env/fonts/LiberationMono-Bold.ttf]:mapping=tex-text;"}{}
  \DeclareFontShape{TU}{pcr}{bx}{n}{<->"[.latex-env/fonts/LiberationMono-Bold.ttf]:mapping=tex-text;"}{}
\fi
\usepackage{microtype}
\usepackage{amsmath,amssymb}
\ifdefined\XeTeXversion
  \DeclareFontShape{OML}{cmm}{m}{it}{%
    <5><6><7>cmmi7 <8><9>cmmi9 <10->cmmi10
  }{}
  \DeclareFontShape{OT1}{cmr}{m}{n}{%
    <5><6>cmr6 <7>cmr7 <8><9>cmr9 <10->cmr10
  }{}
  \DeclareFontShape{OT1}{cmr}{bx}{n}{%
    <5><6><7><8><9>cmbx9 <10->cmbx10
  }{}
  \DeclareFontShape{OT1}{cmr}{b}{n}{%
    <5><6><7><8><9>cmbx9 <10->cmbx10
  }{}
\fi
\usepackage{booktabs}
\usepackage{graphicx}
\usepackage{xcolor}
\ifdefined\XeTeXversion
  \newcommand{\xspace}{\ }
\else
  \usepackage{xspace}
\fi
\usepackage{hyperref}
\usepackage{url}

\newcommand{\method}{\textsc{AcrossWAM1.0}\xspace}

\newcommand{\best}[1]{\textbf{#1}}
\newcommand{\second}[1]{\underline{#1}}

\title{AcrossWAM1.0: A Modular Latent World--Action Stack\\
for Compact Robot Policies}
\author{%
Yafei Zhang$^{1,2}$ \quad Nan Wu$^{1}$\thanks{Corresponding author.}\\
$^{1}$Across Physical AI, Beijing, China\\
$^{2}$Institute of Automation, Chinese Academy of Sciences, Beijing, China\\
\texttt{18453881970@163.com} \quad \texttt{across2026@163.com}
}

\iclrfinalcopy
\begin{document}
\maketitle
\lhead{arXiv preprint}

\begin{abstract}
Latent world--action models avoid rendering future pixels by predicting an action-relevant visual subgoal in feature space. LaWAM established this formulation, but its original presentation left the world model, multimodal backbone, and deployment checkpoint tightly coupled. We introduce \method, a modularization and scaling study of this latent world--action stack. Rather than presenting latent subgoals as a new algorithm, we make the module boundary explicit: a policy adapter produces latent-action and action-generation contexts; a retained latent world decoder grounds the predicted transition in the current scene; and a flow-matching expert generates continuous action chunks. We further separate training-only teachers from the inference graph and provide a verifiable deployment export. On 2,000 paired LIBERO episodes, replacing a Qwen3-VL-2B backbone with Qwen3.5-0.8B yields 97.45\% success versus 98.00\% for the 2B model (a $-0.55$ percentage-point difference; exact McNemar $p=0.266$). This does not prove equivalence, but it meets a prespecified two-point retention criterion. The compact, inference-reachable checkpoint contains 1,472.6M unique parameters, 42.4\% fewer than the original 2B policy, while all retained tensors are bitwise identical to the source checkpoint. Cross-family execution is additionally checked with a MiniCPM-V adapter smoke test; closed-loop cross-family transfer remains an open evaluation. \method therefore contributes an auditable software and evaluation boundary for compact latent world--action policies, distinct from LaWAM's original latent-subgoal contribution.
\end{abstract}

\section{Introduction}
\label{sec:intro}

General-purpose robot manipulation requires a policy to connect semantic intent, visual state, and temporally coherent continuous control. Vision--language--action (VLA) models inherit strong semantic representations from multimodal pretraining~\citep{brohan2023rt2,kim2024openvla,octo2024,black2024pi0}, while world--action models add an explicit prediction of how the scene should change~\citep{cen2025worldvla}. LaWAM showed that this prediction need not be a generated image: an inverse-dynamics latent can be decoded into a spatial visual subgoal and used to condition an action expert~\citep{chen2026lawam}. Its reported operating point combines strong closed-loop success with single-pass latent prediction.

This paper starts from that result rather than claiming it again. We ask a narrower question: \emph{how can the latent world--action computation be made modular, compact, and auditable as a deployable policy stack?} This question matters because replacing a multimodal backbone changes token conventions, hidden dimensions, image-call signatures, checkpoint reachability, and sometimes vocabulary handling. A formulation that is modular on paper can remain tightly coupled in code. Moreover, a smaller backbone is only useful if closed-loop behavior is retained under matched evaluation and if the exported checkpoint actually removes training-only state.

\begin{figure}[t]
    \centering
    \includegraphics[width=\linewidth]{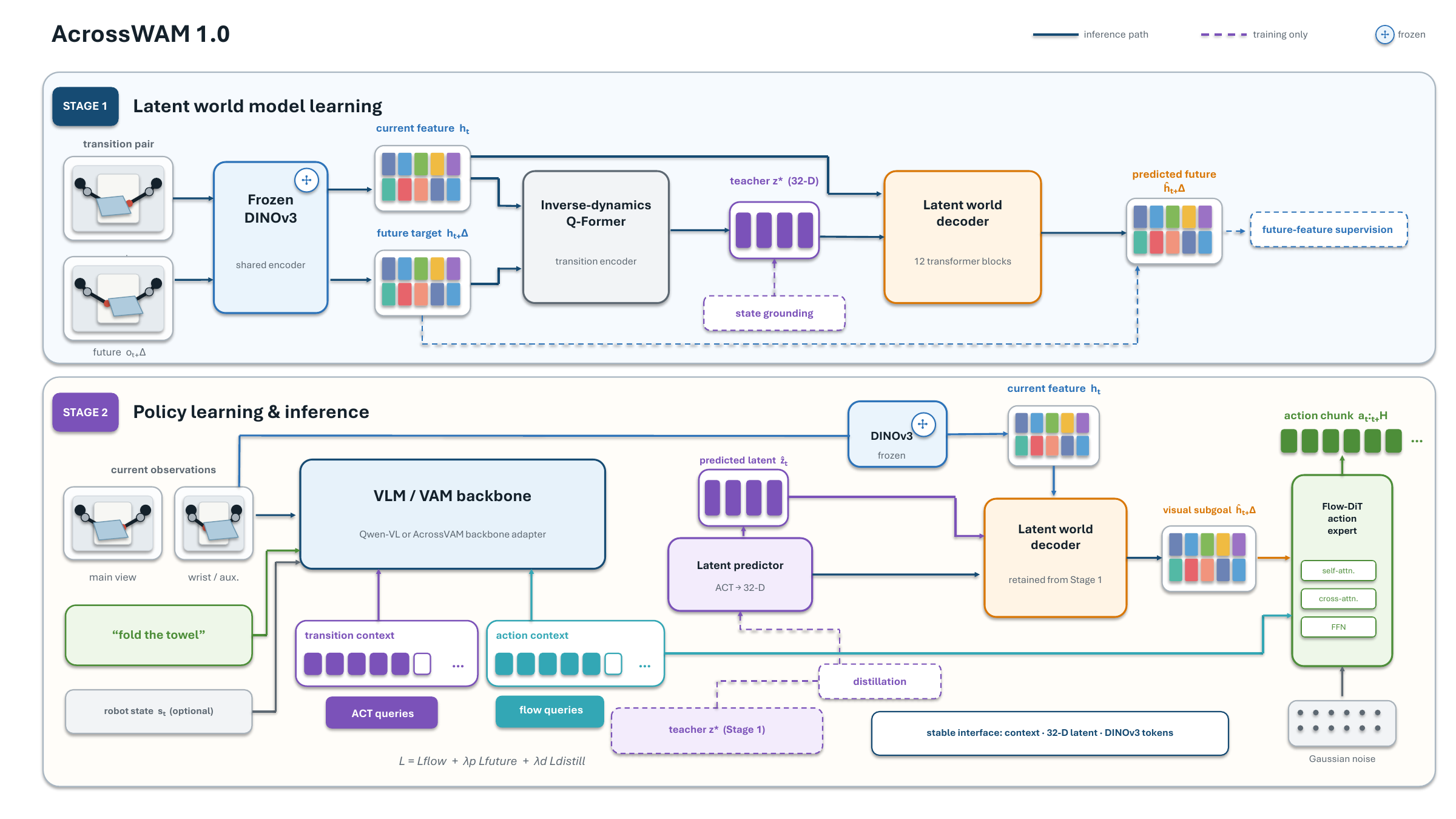}
    \caption{\textbf{AcrossWAM1.0 system boundary.} Stage 1 is inherited from LaWAM: a frozen DINOv3 encoder, inverse-dynamics teacher, and latent world decoder learn an action-bearing transition representation. AcrossWAM1.0 exposes the Stage-2 policy boundary explicitly: a backbone adapter returns separate transition and action contexts, the predicted latent is decoded into a visual subgoal, and a Flow-DiT expert produces an action chunk. Dashed paths and the inverse-dynamics teacher are training-only. In the compact experiment the Stage-1 teacher and DINOv3 are frozen, while the retained decoder is fine-tuned; thus the reported result demonstrates stack modularization and backbone scaling, not frozen cross-family transfer.}
    \label{fig:overview}
\end{figure}

Figure~\ref{fig:overview} shows the resulting boundary. The latent visual subgoal, inverse-dynamics teacher, distillation objective, and flow-conditioned action generation are inherited from LaWAM. AcrossWAM1.0 contributes the surrounding contract and evidence: backbone-specific processing is isolated behind an adapter; transition-query and action-query outputs are returned through a fixed interface; hidden-size alignment is explicit; training-only modules are separated from inference-reachable modules; and compact and full backbones are compared on identical closed-loop episodes. We retain the legacy registry name \texttt{LaWAMFramework} for checkpoint compatibility, but use \method only for this modular stack and its new experiments.

Our contributions are:
\begin{itemize}
    \item We define and implement an explicit policy--world--action contract around the LaWAM computation, including backbone adapters, automatic hidden-size alignment, query ownership, gradient/freeze boundaries, and a training-versus-deployment module split.
    \item We provide a controlled compact-backbone study: on the same 2,000 LIBERO episodes, Qwen3.5-0.8B is 0.55 points below Qwen3-VL-2B with no detected paired difference, while reducing unique deployable parameters by 42.4\%.
    \item We make deployment equivalence auditable by tracing inference-reachable parameters, removing teacher-only modules and duplicate aliases, checking every retained tensor bitwise, and exercising the production loader. We additionally report a cross-family adapter smoke test without conflating it with closed-loop portability.
\end{itemize}

\section{Related Work}
\label{sec:related}

\paragraph{Vision--language--action policies.}
RT-2 introduced the VLA formulation by representing robot actions as tokens in a vision--language model~\citep{brohan2023rt2}. OpenVLA and Octo subsequently provided open, broadly adaptable policy backbones trained on heterogeneous robot data~\citep{kim2024openvla,octo2024,openx2023}. ACT showed that predicting action chunks is effective for high-precision imitation learning~\citep{zhao2023act}, while $\pi_0$ demonstrated that a flow-matching action expert can retain continuous multimodality on top of a pretrained VLM~\citep{black2024pi0}. These approaches provide strong semantic and motor priors, but action generation is not necessarily mediated by an explicit prediction of the action-induced future state.

\paragraph{Latent actions from video.}
Latent action models infer compact transition variables without requiring robot action labels. Genie learns controllable latent actions for interactive environments~\citep{bruce2024genie}; LAPA uses a discrete latent-action bottleneck to pretrain VLAs from videos~\citep{ye2025lapa}; and UniVLA learns task-centric latent actions across embodiments~\citep{bu2025univla}. These works primarily treat latent actions as targets or transferable action abstractions. \method instead emphasizes the forward decoder: the latent action becomes useful to the policy after it is expanded into a spatial future feature grounded in the current observation.

\paragraph{World models for robot control.}
WorldVLA unifies action and image generation autoregressively, illustrating the benefit of coupling world and action prediction~\citep{cen2025worldvla}. Pixel-space WAMs make the future directly inspectable, but their generative path can dominate parameters and latency. LaWAM showed that a latent visual subgoal can preserve the control benefit of future prediction without reconstructing pixels~\citep{chen2026lawam}. \method builds on this result by exposing the policy/world boundary in the implementation and evaluating a compact policy configuration.

\paragraph{From an algorithm to a reusable stack.}
LaWAM answers whether latent visual subgoals can replace pixel futures; AcrossWAM1.0 does not treat that answer as new. Our focus is the unresolved systems and evaluation question around that algorithm: whether module boundaries survive backbone scaling, framework-specific input contracts, and deployment export. This is weaker than universal backbone portability but stronger than a diagram-level claim because it requires matched closed-loop evaluation and tensor-level checkpoint verification. The distinction motivates the evidence levels in Section~\ref{sec:portability}.

\section{Method}
\label{sec:method}

\subsection{Model Identity and Scope}
AcrossWAM1.0 is the complete model specified and evaluated in this paper, with its own modular architecture, trained checkpoints, evaluation protocol, and deployment artifact. It adopts the latent visual-subgoal formulation established by LaWAM~\citep{chen2026lawam} as a technical foundation, but it is neither an alias nor a relabeling of LaWAM. AcrossWAM1.0 independently defines the policy--world--action interface, backbone adapters, hidden-size alignment, query ownership, training/deployment boundary, compact-backbone configuration, and exact export procedure. Accordingly, every result attributed to AcrossWAM1.0 comes from its new checkpoints and evaluations; LaWAM results are cited only as published baselines. This separates the model identity and empirical claims of AcrossWAM1.0 while keeping its algorithmic lineage explicit.

\subsection{Problem Formulation}
Let $o_t$ be the current multi-view visual observation, $\ell$ a language instruction, $s_t$ an optional proprioceptive state, and $\mathbf{a}_{t:t+H-1}\in\mathbb{R}^{H\times d_a}$ an action chunk spanning physical horizon $\Delta$. A standard VLA directly parameterizes $p(\mathbf{a}_{t:t+H-1}\mid o_t,\ell,s_t)$. We introduce a latent action $z_t\in\mathbb{R}^{d_z}$ and a latent visual subgoal $\hat{h}_{t+\Delta}$, yielding
\begin{equation}
\label{eq:factorization}
    p(\mathbf{a},\hat{h}_{t+\Delta},z_t\mid o_t,\ell,s_t)
    =p_\theta(z_t\mid o_t,\ell)\,
     p_\omega(\hat{h}_{t+\Delta}\mid h_t,z_t)\,
     p_\eta(\mathbf{a}\mid c_\theta,h_t,\hat{h}_{t+\Delta},s_t),
\end{equation}
where $h_t=f_\psi(o_t)$ is a frozen DINOv3 patch representation, $c_\theta$ denotes the multimodal backbone context, $p_\omega$ is the latent world decoder, and $p_\eta$ is the action expert. Equation~\ref{eq:factorization} makes the predicted scene change an explicit intermediate variable rather than an implicit side effect of the policy hidden state.

\subsection{Stage 1: Learning an Action-Bearing Latent World Model}

\paragraph{Frozen feature space.}
For each transition pair $(o_t,o_{t+\Delta})$, a frozen DINOv3 encoder~\citep{simeoni2025dinov3} produces patch tokens $h_t=f_\psi(o_t)$ and $h_{t+\Delta}=f_\psi(o_{t+\Delta})$, with $h\in\mathbb{R}^{K\times d_v}$. Freezing $f_\psi$ stabilizes the target space and prevents the transition objective from collapsing by jointly moving both predictions and targets. Dense pretrained tokens preserve object and geometry information while avoiding pixel reconstruction.

\paragraph{Inverse dynamics and latent action.}
An inverse-dynamics Q-Former $q_\phi$ attends to the temporally positioned token pair and returns one low-dimensional posterior latent,
\begin{equation}
    q_\phi(z_t\mid h_t,h_{t+\Delta})
    =\mathcal{N}\!\left(\mu_\phi,\operatorname{diag}(\sigma_\phi^2)\right),
    \quad z_t=\mu_\phi+\sigma_\phi\odot\epsilon,\quad \epsilon\sim\mathcal{N}(0,I).
\end{equation}
In the released configuration, $d_z=32$ and the encoder uses one query. Because the bottleneck observes both endpoints but has limited capacity, it is encouraged to encode the transition that distinguishes them rather than the appearance shared by both frames.

\paragraph{Latent world decoding.}
The decoder $D_\omega$ receives current tokens and the latent action. Each transformer block uses adaptive LayerNorm modulation derived from $z_t$, followed by self-attention and an MLP:
\begin{align}
    \bar h^{(i)} &= \operatorname{AdaLN}\!\left(h^{(i)};z_t\right),\\
    h^{(i+1)} &= h^{(i)} + g^{(i)}_{\mathrm{attn}}(z_t)\operatorname{SA}(\bar h^{(i)})
       +g^{(i)}_{\mathrm{mlp}}(z_t)\operatorname{MLP}(\bar h^{(i)}).
\end{align}
The output $\tilde h_{t+\Delta}=D_\omega(h_t,z_t)$ is trained to match the frozen future feature. Action-conditioned modulation is appropriate because a global transition variable must induce spatially different changes across patch tokens.

\paragraph{State grounding and stage-one objective.}
For robot samples with valid state masks, a category-specific predictor $g_\xi$ maps $(s_t,z_t)$ to the state displacement $\Delta s_t$. The complete loss is
\begin{equation}
\label{eq:stage1}
    \mathcal{L}_{\mathrm{S1}}
    =\underbrace{\operatorname{SmoothL1}_{\beta=0.1}(\tilde h_{t+\Delta},h_{t+\Delta})}_{\mathcal{L}_{\mathrm{future}}}
    +\lambda_s\underbrace{\left\|m_s\odot(g_\xi(s_t,z_t)-\Delta s_t)\right\|_2^2}_{\mathcal{L}_{\mathrm{state}}}
    +\beta_{\mathrm{KL}}D_{\mathrm{KL}}(q_\phi\|\mathcal{N}(0,I)).
\end{equation}
Here $m_s$ masks padded or unavailable state dimensions. Future-feature supervision makes $z_t$ sufficient for visual dynamics; state grounding discourages it from representing only nuisance appearance changes; and the KL term produces a regular latent space that the policy can later imitate. After stage 1, $f_\psi$ and $D_\omega$ form the latent world module, while $q_\phi$ becomes a training-only teacher. This procedure is inherited from LaWAM; it is included to make the complete computation self-contained.

\subsection{Stage 2: Predicting Latent Actions from Language and Vision}

\paragraph{Backbone and query layout.}
The policy backbone receives the instruction, main-view image, optional wrist/auxiliary views, and learned placeholder embeddings. Two disjoint query groups have different ownership: eight \emph{latent-action queries} (called ACT queries in the code and Fig.~\ref{fig:overview}) collect the multimodal context required to infer the transition, while eight \emph{flow queries} expose context to the continuous action expert. An adapter must return contextual tokens, the two query slices, and the image-processing arguments expected by its backbone. The reported closed-loop models use Qwen-family backbones~\citep{bai2025qwen3vl}; the more general VLM/VAM label denotes an interface target, not a completed cross-family benchmark.

The hidden states at the latent-action positions are compressed by a one-query cross-attention mapper $M_\rho$ into the predicted latent action $\hat z_t=M_\rho(c_\theta^{\mathrm{LA}})$. The frozen stage-one teacher extracts $z_t^*=q_\phi(h_t,h_{t+\Delta})$ from the observed transition. We align the policy prior with this target using
\begin{equation}
    \mathcal{L}_{\mathrm{distill}}=\|\hat z_t-z_t^*\|_2^2.
\end{equation}
This loss assigns the latent-action queries a specific dynamic meaning rather than allowing them to become arbitrary action-head features.

\paragraph{Latent visual subgoal.}
The policy-driven latent is decoded with the retained world model,
\begin{equation}
    \hat h_{t+\Delta}=D_\omega(h_t,\hat z_t),\qquad
    \mathcal{L}_{\mathrm{subgoal}}=\|\hat h_{t+\Delta}-h_{t+\Delta}\|_2^2.
\end{equation}
The predicted subgoal has the same spatial token structure as the current DINOv3 feature. Unlike a single latent action vector, it can specify which image regions should change and how the predicted transition is grounded in the current embodiment and scene.

\subsection{Subgoal-Conditioned Flow-Matching Action Expert}
The action expert uses flow matching~\citep{lipman2023flow} in a conditional diffusion transformer (DiT)~\citep{peebles2023dit}. One conditioning stream carries the backbone context, including flow-query states; a second stream carries the concatenated current and predicted-future visual tokens $(h_t,\hat h_{t+\Delta})$. Alternating self- and cross-attention blocks fuse semantic intent with the predicted dynamics.

For a ground-truth normalized action chunk $\mathbf{a}_1$ and Gaussian noise $\mathbf{a}_0\sim\mathcal{N}(0,I)$, we sample time $\tau\in[0,1]$ and form
\begin{equation}
    \mathbf{a}_\tau=(1-\tau)\mathbf{a}_0+\tau\mathbf{a}_1,
    \qquad u_\tau=\mathbf{a}_1-\mathbf{a}_0.
\end{equation}
The expert $v_\eta$ regresses this conditional vector field:
\begin{equation}
\label{eq:flow}
    \mathcal{L}_{\mathrm{flow}}
    =\mathbb{E}_{\tau,\mathbf{a}_0,\mathbf{a}_1}
      \left[\left\|v_\eta(\mathbf{a}_\tau,\tau\mid c_\theta,h_t,\hat h_{t+\Delta},s_t)-u_\tau\right\|_2^2\right].
\end{equation}
At test time, an ODE solver integrates the learned field from Gaussian noise to an action chunk. The released configuration predicts $H=50$ actions over 1.2\,s and uses 10 integration steps.

The stage-two loss is
\begin{equation}
\label{eq:stage2}
    \mathcal{L}_{\mathrm{S2}}
    =\mathcal{L}_{\mathrm{flow}}
     +\lambda_p\mathcal{L}_{\mathrm{subgoal}}
     +\lambda_d\mathcal{L}_{\mathrm{distill}},
\end{equation}
with $\lambda_p=\lambda_d=0.1$ in the compact Qwen configuration. Future targets are detached from the action expert when configured, preventing an easy ground-truth future path from bypassing the learned policy prior.

\subsection{Inference, Module Boundaries, and Evidence Levels}
\label{sec:portability}
Inference uses only current observations and an instruction. The backbone predicts $\hat z_t$ through latent-action queries; $D_\omega$ produces $\hat h_{t+\Delta}$ in one pass; and the action expert integrates from noise using the flow-query context and latent subgoal. The inverse-dynamics encoder, state-grounding head, and future frame are absent from the inference graph.

We distinguish three claims that are often collapsed under ``portability.'' \textbf{Interface compatibility} means that an adapter can execute the required input/output contract; it is verified for two Qwen variants and by a MiniCPM-V forward/backward smoke test. \textbf{Closed-loop backbone scaling} means that a newly trained policy using the same module contract retains task performance; it is evaluated for Qwen3-VL-2B and Qwen3.5-0.8B. \textbf{Frozen cross-family transfer} would require sharing an unchanged $D_\omega$ (or a precisely bounded adapter) across structurally different backbones and then evaluating control. We do not claim the third level: in the reported compact run, DINOv3 and the teacher are frozen but the latent world decoder, policy backbone, mapper, and action expert are optimized. Thus the current result establishes a reusable implementation boundary and compact scaling within the Qwen family, not a universal frozen world interface.

\section{Experiments}
\label{sec:experiments}
Our experiments target the claims that are new to AcrossWAM1.0: \textbf{(Q1)} how much closed-loop performance is retained by a compact backbone, \textbf{(Q2)} how much inference-reachable state can be removed without changing the policy tensors, and \textbf{(Q3)} which levels of adapter portability are currently supported. LaWAM's published 98.6\% LIBERO, 92.64\%/89.80\% RoboTwin, 90.0\% real-robot, and 187\,ms latency results are inherited reference points~\citep{chen2026lawam}; they are not relabeled as AcrossWAM1.0 results.

\subsection{Experimental Setup}
\paragraph{Models and training.}
The comparison uses the same latent world--action stack with either a Qwen3-VL-2B or Qwen3.5-0.8B policy backbone. The compact model is trained for 25,000 steps; its final training loss is 0.01428 and validation MSE is 0.002313. DINOv3 feature extraction and the inverse-dynamics teacher are frozen, whereas the VLM, latent mapper, world decoder, and flow expert are optimized. This detail is essential: the experiment measures backbone scaling of the modular stack, not transfer through a frozen world decoder.

\paragraph{Benchmarks.}
LIBERO contains long-horizon, object, goal, and spatial suites~\citep{liu2023libero}. We run 50 trials for each of 10 tasks per suite, for 500 episodes per suite and 2,000 episodes per policy. Pairing keys are (suite, task, episode index); both policies use the same seed, task order, and episode initialization, with no missing pairs.

\paragraph{Statistics.}
Our prespecified engineering retention criterion is a drop of no more than two percentage points. We additionally use an exact paired McNemar test and report discordant counts. A nonsignificant test is not evidence of equivalence, and one checkpoint per backbone does not measure training-seed variance.

\subsection{Recent Benchmark Context}
Table~\ref{tab:benchmarks} places the compact checkpoint beside recent VLA and world--action baselines. We prioritize methods released from 2024 onward and include both standard LIBERO and RoboTwin 2.0~\citep{chen2025robotwin} when a compatible aggregate is available. Success rates and parameter counts are transcribed from the method papers and their recent benchmark tables~\citep{black2024pi0,kim2025openvlaoft,pi05_2025,bu2025univla,goyal2025vla0,sun2026vlajepa,guo2026priorvla,bi2025motus,yuan2026fastwam,kim2026cosmospolicy,li2026causalworld,chen2026lawam,pan2026selfwam,ma2026fasterwamdot,zhao2026fasterwamsparse}, not from our reproductions; we leave the parameter entry blank when a comparable total is unavailable rather than imputing it. Training recipes, embodied pretraining, evaluation implementations, and parameter-accounting conventions differ, so the table provides external context rather than a controlled leaderboard. PriorVLA evaluates a different 13-task Easy/Hard RoboTwin protocol and is therefore left blank in the RoboTwin-50 columns~\citep{guo2026priorvla}. Two independent August 2026 papers use the name Faster-WAM; we disambiguate them by their DoT and SparseMoT designs.

\begin{table*}[t]
\centering
\caption{\textbf{Model size and success rates (\%) reported by recent robot-policy papers.} Params is the source-reported model size in billions; conventions vary, and some pixel WAM sources exclude the VAE or text encoder. AcrossWAM1.0 uses the exact unique inference-reachable count of its Qwen3.5-0.8B deployment checkpoint. Release denotes the first public preprint year and can precede the formal proceedings year. LIBERO is the average over the four standard suites. RoboTwin 2.0 reports the 50-task Clean/Randomized setting with 100 trials per task when available. ``--'' means that the cited paper does not report a comparable value.}
\label{tab:benchmarks}
\small
\setlength{\tabcolsep}{3.4pt}
\begin{tabular}{llcrrrr}
\toprule
Method & Family & Release & Params (B) & LIBERO & \multicolumn{2}{c}{RoboTwin 2.0} \\
\cmidrule(lr){6-7}
 & & & & Avg. & Clean & Random. \\
\midrule
$\pi_0$ & VLA & 2024 & 3.3 & 94.1 & 65.92 & 58.40 \\
OpenVLA-OFT & VLA & 2025 & 7.0 & 97.1 & -- & -- \\
$\pi_{0.5}$ & VLA & 2025 & 3.5 & 96.9 & 82.74 & 76.76 \\
UniVLA & latent-action VLA & 2025 & 7.0 & 95.2 & -- & -- \\
VLA-0 & VLA & 2025 & 3.0 & 94.7 & -- & -- \\
Motus & joint latent WAM & 2025 & 8.0 & 97.7 & 88.66 & 87.02 \\
\midrule
VLA-JEPA & latent WAM & 2026 & 3.0 & 97.2 & -- & -- \\
Fast-WAM & training-time WAM & 2026 & 6.0 & 97.6 & 91.42 & 91.86 \\
Cosmos-Policy & pixel WAM & 2026 & 2.1 & 98.5 & -- & -- \\
LingBot-VA & pixel WAM & 2026 & 5.3 & 98.5 & \best{92.90} & 91.50 \\
LaWAM & latent WAM & 2026 & 2.3 & 98.6 & 92.64 & 89.80 \\
PriorVLA & VLA adaptation & 2026 & -- & \best{99.1} & -- & -- \\
SelfWAM & self-grounded WAM & 2026 & -- & -- & 92.16 & \best{93.08} \\
Faster-WAM (DoT) & efficient WAM & 2026 & -- & 98.5 & 89.70 & 88.64 \\
Faster-WAM (SparseMoT) & efficient WAM & 2026 & -- & \second{99.0} & \second{92.80} & \second{92.30} \\
\midrule
\textbf{AcrossWAM1.0 (ours)} & modular latent WAM & 2026 & 1.5 & 97.45 & 92.09 & 89.12 \\
\bottomrule
\end{tabular}
\end{table*}

AcrossWAM1.0 reaches 97.45\% on standard LIBERO. Among the cited results, it is 0.35 points above OpenVLA-OFT, 0.25 above VLA-JEPA, and 0.55 above $\pi_{0.5}$, while trailing the newest 98.5--99.1\% reports. This is not a state-of-the-art claim. The result instead supports the paper's compactness question: it is obtained with a 1,472.6M-parameter inference-reachable checkpoint and is backed by a paired comparison against the 2B instantiation below. We do not fill the RoboTwin columns with inherited LaWAM numbers or with the preliminary language diagnostic because neither is an AcrossWAM1.0 benchmark evaluation.

\subsection{Compact-Backbone Study}
\begin{table*}[t]
\centering
\caption{\textbf{Paired LIBERO backbone comparison (success, \%).} Both checkpoints use the same 2,000 episode keys. Deployable parameters count unique inference-reachable parameters after export.}
\label{tab:compact}
\small
\begin{tabular}{lrrrrrr}
\toprule
Backbone & Long & Object & Goal & Spatial & Pooled & Deploy params\\
\midrule
Qwen3-VL-2B & 96.6 & 99.2 & 96.8 & 99.4 & 98.00 (1960/2000) & 2,555.2M\\
Qwen3.5-0.8B & 95.0 & 99.4 & 96.8 & 98.6 & 97.45 (1949/2000) & 1,472.6M\\
\bottomrule
\end{tabular}
\end{table*}

The compact model is 0.55 percentage points lower overall and satisfies the two-point retention criterion. The discordant outcomes are 46 episodes won only by the 2B policy and 35 won only by the 0.8B policy, yielding exact McNemar $p=0.2664$. We therefore conclude that this evaluation did not detect a paired difference; we do not claim statistical equivalence or non-inferiority across training seeds. The result is nevertheless operationally meaningful: the same latent world--action module contract accommodates a substantially smaller backbone and retains 1,949 successful closed-loop episodes.

\subsection{Qualitative LIBERO Prediction Visualization}
\begin{figure*}[t]
  \centering
  \includegraphics[width=0.94\textwidth]{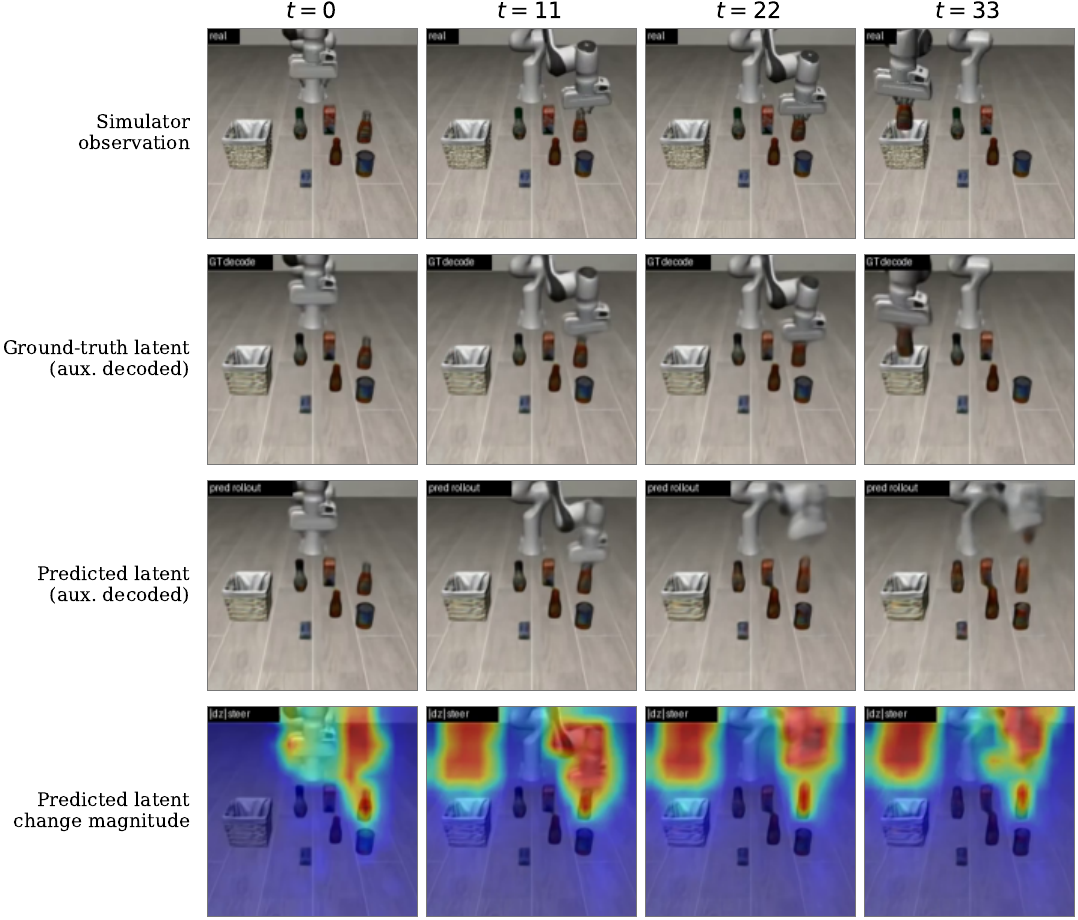}
  \caption{\textbf{Qualitative latent prediction on a held-out LIBERO simulator trajectory.} Columns show rollout steps 0, 11, 22, and 33. Rows show the simulator observation, an auxiliary RGB rendering of the encoded ground-truth latent, an auxiliary RGB rendering of the autoregressively predicted latent, and the spatial magnitude of the predicted latent change. The predicted sequence preserves the scene layout and task-relevant motion at shorter horizons, while blur accumulates under long autoregressive rollout. The auxiliary decoder is used only for this visualization and is absent from policy inference and evaluation.}
  \label{fig:libero-prediction}
\end{figure*}

Figure~\ref{fig:libero-prediction} makes a held-out LIBERO simulator rollout in the learned latent space visually inspectable. Starting from the initial latent state, the diagnostic autoregressively predicts 33 steps and compares selected horizons with encoded ground-truth states. An auxiliary RGB decoder renders both latent sequences solely for interpretation; it is not an input to the policy, does not alter the predicted actions, and is not used to compute the success rates in Table~\ref{tab:benchmarks}. The visualization is therefore qualitative evidence about the latent prediction interface rather than a pixel-generation claim.

\subsection{Deployment Reachability and Exact Export}
\begin{table}[t]
\centering
\caption{\textbf{Unique parameter counts across compacting and export.} ``Raw'' includes training-only modules; ``deploy'' retains only inference-reachable state.}
\label{tab:deploy}
\small
\begin{tabular}{lrr}
\toprule
Artifact & Unique parameters & Change from 2B raw\\
\midrule
Qwen3-VL-2B raw policy & 2,555.2M & --\\
Qwen3.5-0.8B raw policy & 1,883.8M & $-26.3\%$\\
Qwen3.5-0.8B deploy export & 1,472.6M & $-42.4\%$\\
\bottomrule
\end{tabular}
\end{table}

The export removes the inverse-dynamics teacher, state decoder, and unused vector-quantization state, while retaining the world decoder used at inference. It also deduplicates aliases caused by registering the same flow and VLM modules through multiple paths. Verification accounts for every source key: 1,078 retained tensors are bitwise identical, 321 training-only keys (411.1M parameters) are deliberately removed, and 718 removed alias keys are bitwise reproducible from retained keys. The compact checkpoint materializes successfully through the production CPU loader with the VLM, latent mapper, world decoder, latent/action query parameters, and flow head present. This establishes checkpoint-state equivalence for the reachable graph; a full accelerator-and-simulator smoke test of the exported file is not yet included.

\subsection{Adapter Contract Validation}
The Qwen3-VL-2B and Qwen3.5-0.8B results establish closed-loop scaling within one backbone family. To probe framework coupling, we also implement the same call contract for MiniCPM-V 4.6. A forward/backward smoke test verifies that \texttt{inputs\_embeds} and \texttt{pixel\_values} can be used together, injected query tokens change the output (mean absolute hidden-state change 0.33), image arguments no longer depend on Qwen-specific \texttt{image\_grid\_thw}, the 1,024-D hidden size is aligned automatically, and the training loss decreases. This is useful evidence for the adapter abstraction, but it is not a manipulation result. MiniCPM-V also exposes a portability failure mode: unlike the Qwen checkpoint, its vocabulary has no unused embedding rows for placeholder queries, so resizing and freezing rules must be handled explicitly.

\subsection{Language Grounding Diagnostic}
As a secondary diagnostic, we change only \emph{left} versus \emph{right} in a controlled RoboTwin instruction and record where the object is placed. Among completed conflicting-instruction placements, the legacy policy follows the stated direction in 52/190 cases (27.4\%). A language classifier-free-guidance variant reaches 93/185 (50.3\%; Fisher exact $p=1.0\times10^{-5}$), while conflicting-instruction collapse is unchanged (36.7\% versus 38.3\%, $p=0.74$). On non-conflicting instructions, however, collapse increases from 1\% to 10\%. We report this as a diagnostic and limitation, not as the core AcrossWAM1.0 contribution: aggregate success can hide weak language grounding, and stronger guidance currently trades adherence against task completion.

\section{Limitations}
\label{sec:limitations}
AcrossWAM1.0 is a modularization, compact-scaling, and deployment study built on LaWAM; it is not a new latent-subgoal learning algorithm. Its closed-loop evidence currently covers two sizes in the Qwen family and LIBERO only. The MiniCPM-V result tests an execution contract, not control quality, and no cross-family AcrossWAM1.0 closed-loop result is reported. Moreover, the compact experiment fine-tunes the latent world decoder, so it does not establish that one frozen decoder transfers across heterogeneous backbones. Each backbone is represented by one trained checkpoint: episode pairing controls evaluation stochasticity but not training-seed variation. Compact-model latency, peak memory, and accelerator/simulator validation of the exported checkpoint are also missing. Finally, the left/right diagnostic covers only two RoboTwin tasks and exposes a real adherence--completion trade-off; it should not be generalized to language grounding as a whole.

\section{Conclusion}
\label{sec:conclusion}
We presented AcrossWAM1.0, an auditable modular stack around LaWAM's latent world--action computation. Its contribution is not the latent visual subgoal itself, but an explicit backbone/world/action contract, a controlled compact-backbone evaluation, and an exact deployment export. A Qwen3.5-0.8B policy reaches 97.45\% on 2,000 LIBERO episodes versus 98.00\% for Qwen3-VL-2B; the paired test detects no difference, while the exported compact policy reduces unique inference-reachable parameters by 42.4\%. The evidence supports compact scaling within the Qwen family and cross-family adapter execution. It does not yet support frozen, performance-preserving portability across arbitrary policy backbones. This separation between demonstrated results and stronger future hypotheses is the central design principle of AcrossWAM1.0.

\subsection*{Ethics statement}
AcrossWAM1.0 is intended for research on robot manipulation. Deployment on physical hardware can cause collisions or damage if the policy encounters out-of-distribution scenes, ambiguous language, or sensor failures. Real-robot use should include workspace limits, collision monitoring, emergency stops, and human supervision. The datasets used by this project are public robot or egocentric-video resources; downstream users remain responsible for their licenses, privacy conditions, and safe operating procedures.

\subsection*{Reproducibility statement}
The accompanying repository contains model code, stage-one and stage-two configurations, data adapters, training scripts, benchmark evaluators, paired statistical tests, and deployment-checkpoint verification. Section~\ref{sec:method} specifies the factorization, ownership boundaries, and objectives; Section~\ref{sec:experiments} states the episode keys, sample counts, statistical test, parameter accounting, and limits of each validation level. Machine-readable paired and export-verification results accompany the code.

\subsection*{AI use statement}
Generative AI tools were used to organize an initial manuscript draft, improve wording, and perform consistency checks against the project code and machine-readable results. The authors reviewed the generated text, verified quantitative claims against project artifacts, checked bibliographic metadata against primary papers, and take responsibility for the final content. No AI-generated experimental result is reported.

\bibliography{iclr2027_conference}
\bibliographystyle{iclr2027_conference}

\appendix
\section{Implementation Details}
\label{app:implementation}
The compact Qwen configuration uses a 1,024-D backbone hidden state, 32-D latent action, eight latent-action queries, eight flow queries, 50-step action horizon, 1.2\,s physical chunk duration, and 10 flow integration steps. The DiT action expert has hidden size 1,024, 16 attention heads, and 16 layers. Stage-two training uses $\lambda_p=\lambda_d=0.1$, repeated noise sampling twice per batch, gradient clipping at 1.0, AdamW with peak learning rate $10^{-4}$ for the VLM and flow expert, and $3\times10^{-4}$ for the latent world decoder. DINOv3 feature extraction and the stage-one teacher are frozen during policy training; the decoder is trainable.

\section{Compact LIBERO Results by Suite}
\begin{table}[h]
\centering
\caption{Per-suite paired backbone results (500 episodes per suite).}
\small
\begin{tabular}{lrrrr}
\toprule
Suite & Qwen3-VL-2B & Qwen3.5-0.8B & $\Delta$ (pp) & $p$\\
\midrule
LIBERO-Long & 96.6 & 95.0 & -1.6 & 0.243\\
LIBERO-Object & 99.2 & 99.4 & +0.2 & 1.000\\
LIBERO-Goal & 96.8 & 96.8 & 0.0 & 1.000\\
LIBERO-Spatial & 99.4 & 98.6 & -0.8 & 0.344\\
\midrule
Pooled & 98.00 & 97.45 & -0.55 & 0.266\\
\bottomrule
\end{tabular}
\end{table}

\section{Evidence Taxonomy and Explicit Non-Claims}
\label{app:evidence}
For clarity, the paper uses \emph{verified} only for results backed by completed artifacts. Closed-loop scaling is verified for Qwen3-VL-2B versus Qwen3.5-0.8B on LIBERO. Exact deployment export is verified at the state-dictionary and CPU-loader levels. Cross-family adapter execution is verified by the MiniCPM-V forward/backward smoke test. The following are explicitly not claimed: statistical equivalence across training seeds; closed-loop MiniCPM-V or cross-family AcrossWAM1.0 performance; a decoder that remains frozen across backbone families; compact-model latency or peak-memory improvement; and new ownership of LaWAM's published algorithm or benchmark results.

\end{document}